\documentclass[11pt,a4paper]{article}
\usepackage[]{naacl2021}
\usepackage{times}
\usepackage{url}
\usepackage{latexsym}
\usepackage{graphicx, subfig}
\usepackage{multirow}
\usepackage{amssymb}
\usepackage{enumerate}
\usepackage{enumitem}
\usepackage{mathrsfs}
\usepackage{amsmath}
\usepackage{subfig}
\usepackage{tabu}
\usepackage{color}

\definecolor{Ao}{rgb}{0.0, 0.5, 0.0}
\definecolor{cornellred}{rgb}{0.7, 0.11, 0.11}
\usepackage{microtype}

\title{Inflected Forms Are Redundant in Question Generation Models}

\author{Xingwu Sun \\
  University of Macau \\
  sunxingwu01@gmail.com \\\And

  Hongyin Tang \\
  Meituan Inc. \\
  tanghongyin@meituan.com \\\And
  
  Chengzhong Xu \\
  University of Macau \\
  czxu@um.edu.mo \\}

\date{}

\begin{document}
\maketitle
\begin{abstract}
Neural models with an encoder-decoder framework provide a feasible solution to Question Generation (QG). However, after analyzing the model vocabulary we find that current models (both RNN-based and pre-training based) have more than 23\% inflected forms. As a result, the encoder will generate separate embeddings for the inflected forms, leading to a waste of training data and parameters. Even worse, in decoding these models are vulnerable to irrelevant noise and they suffer from high computational costs. In this paper, we propose an approach to enhance the performance of QG by fusing word transformation. Firstly, we identify the inflected forms of words from the input of encoder, and replace them with the root words, letting the encoder pay more attention to the repetitive root words. Secondly, we propose to adapt QG as a combination of the following actions in the encode-decoder framework: generating a question word, copying a word from the source sequence or generating a word transformation type. Such extension can greatly decrease the size of predicted words in the decoder as well as noise. We apply our approach to a typical RNN-based model and \textsc{UniLM} to get the improved versions. We conduct extensive experiments on SQuAD and MS MARCO datasets. The experimental results show that the improved versions can significantly outperform the corresponding baselines in terms of BLEU, ROUGE-L and METEOR as well as time cost.
%Secondly, we propose to extend the encoder-decoder framework with a three-way soft switch, where the switch decides the mode and corresponding action of the decoder: 
\end{abstract}

\section{Introduction}
Question Generation (QG) aims to generate natural language questions from given text. It can aid several applications: (1) QG can help create educational materials by generating questions for reading comprehension materials~\cite{heilman2010good,du2017learning}. (2) It can be used to automatically curate question answering datasets~\cite{duan2017question}. (3) It can also aid dialogue systems by actively asking meaningful questions. Typically, QG includes two sub-tasks: (1) determine the targets that should be asked (e.g., sentences, phrases or words), and (2) produce the surface-form of the question. In this work, we focus on the sub-task of surface-form generation of questions by assuming that the targets are given.

\begin{figure}
    \centering
    \includegraphics[width=0.5\textwidth]{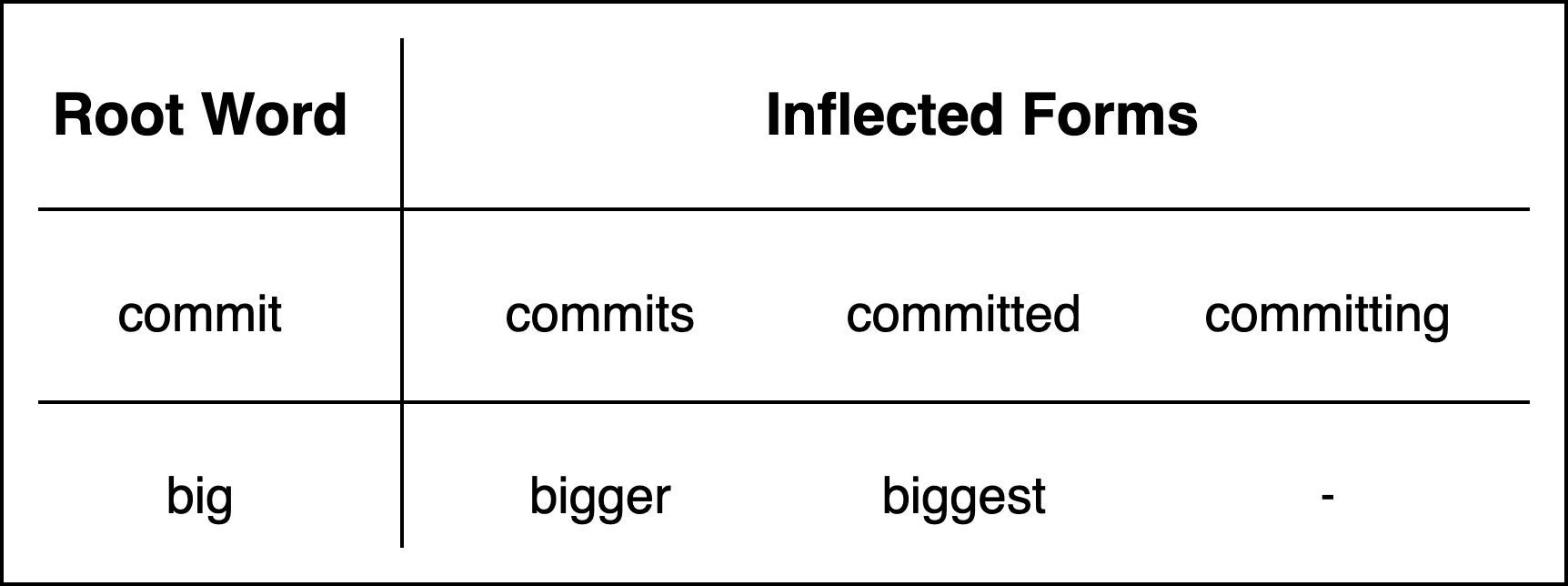}
    \vspace{-0.1cm}
    \caption{Two root words and their corresponding inflected forms in the vocabulary of QG models.}
    \label{fig:commit}
    \vspace{-0.5cm}
\end{figure}

Recent neural QG models, e.g., RNN-based sequence-to-sequence models and pre-training based models, have made great progress in generating proper questions. In particular, \textsc{UniLM}~\cite{dong2019unified}, which applies a pre-training model for QG, is the state-of-the-art model with the best performance. However, those models all suffer high computational costs in decoding. We carefully examine existing neural QG models which have an encoder-decoder architecture in general and find that existing models remain two issues. Firstly, in the encoder, the root word and its inflected forms, e.g., “commit” and its inflected forms “committed” and “committing” as shown in Figure \ref{fig:commit}, might all occur in the vocabulary of the encoder. According to our statistics, in the widely used QG dataset SQuAD~\cite{rajpurkar2016squad}, the top-$10000$ frequent words contain 3718 inflected forms. Moreover, even though \textsc{UniLM} applies the WordPiece segmentation to reduce the vocabulary size, its vocabulary still contains about 23.18\% (6722/28996) inflected forms. As a result, the encoder will generate individual embeddings for these words, which seems a little redundant for QG. Even worse, they occupy the space in the vocabulary of encoder. As a result, the encoder will generate separate embeddings for the inflected words, leading to a waste of training data and parameters. Secondly, in the decoder, the same large set of words are employed to generate a question regardless of the input. Therefore, these models are vulnerable to irrelevant noise. Besides, while decoding each word of a question, probability distribution of words in the entire static and fairly large vocabulary have to be calculated, which slows the decoding process down. 

To decrease the computational cost and improve the performance of QG, we propose an approach named as word transformation approach. The approach is inspired by the feature of the inflected language and our daily reading experience. The former refers to the fact that words with different inflected forms may divert the focus to QG. The latter refers to our observations that most words in the generated question can be copied or simply transformed from the words in the original source sequence, except for the question word. Therefore, in our approach, we treat QG as a combination of word transformation, word copying and question word generation. In detail, in the question word generation, a question word is generated from a limited question word vocabulary. In the transformation type generation, an inflected form type is generated from a small word transformation type vocabulary. For example, type ``\#\#ed" may be generated and will be used for transforming a verb to its past tense in the final step. In the word copying, words are copied from the source sequence. Besides, we identify the inflected forms of words from the input of encoder, and replace them with their root forms, letting the encoder pay more attention to the repetitive root words and take in more different words for training. We have applied our approach to a typical RNN-based model and a pre-trained transformer-based model \textsc{UniLM} and get the corresponding improved versions.
%for facilitating the decoder to generate the question in an effective and efficiency way, we propose a three-way soft switch, which decides the mode the decoder will enter and the action the decoder will take in the next time step. The modes of decoder are question word generation mode, transformation type generation mode and copy mode, respectively. 

The main contributions of this paper can be summarized as follows.
\begin{itemize}
\setlength{\parsep}{0pt}
\setlength{\parskip}{0pt}
\item We find that inflected words take up unnecessary spaces in vocabulary and define a series of word transformation types to facilitate transforming a word into its inflected forms.
%We treat QG as a combination of word transformation, word copying and question word generation and define a series of word transformation types to facilitate transforming a word into its inflected form. 
\item We only keep the root words in vocabulary of the encoder to make the best of the training data and the limited vocabulary space, which is also applicable in many other NLP models.
%\item We propose a three-mode decoding for QG, i.e., question word generation mode, transformation type generation mode and copy mode to avoid generating each word from a large vocabulary and enhance the performance. 
\item We propose to simplify the decoding process of QG by question word generation, transformation type generation and word copying, to avoid generating each word from a large vocabulary and enhance the performance.
\item For evaluating the effectiveness and efficiency of our proposed approach, we conduct extensive experiments on two large scale datasets, i.e., SQuAD and MS MARCO datasets and compare the baselines with the corresponding improved versions. The experimental results show that the improved versions can significantly outperform the baselines. % including \textsc{UniLM}.
%the current state-of-the-art model, i.e.,
\end{itemize}

\section{Related Work}
\begin{table*}[!t]
\centering
\begin{tabular}{|c|c|l|}
\hline
\textbf{
Part-of-speech}                & \multicolumn{1}{c|}{\textbf{Transformation Types}} & \multicolumn{1}{c|}{\textbf{Transformation Type Description}}          \\ \hline
\multirow{4}{*}{verb}  
              & \#\#ing  & converting the verb to its present participle    \\ \cline{2-3} 
              & \#\#vs   & converting the verb to its singular present \\ \cline{2-3} 
              & \#\#ed & converting the verb to its past tense     \\ 
              \cline{2-3} 
              & \#\#edp & converting the verb to its past participle             \\ \hline
noun & \#\#ns   & converting the noun to its plural word      \\ \hline
\multirow{2}{*}{adjective} & \#\#jer                                  & converting the adjective to its comparative form \\ \cline{2-3} 
              & \#\#jest & converting the adjective to its superlative form               \\ \hline
\multirow{2}{*}{adverb} & \#\#ver                                  & converting the adverb to its comparative form \\ \cline{2-3} 
              & \#\#vest & converting the adverb to its superlative form               \\ \hline              
\end{tabular}
\vspace{-0.1cm}
\caption{Word transformation types and the corresponding descriptions.}
\label{transformation}
\vspace{-0.5cm}
\end{table*}
The existing QG approaches can be broadly classified into two kinds: rule-based and neural network-based. The rule-based approaches rely on hand-crafted rules, while the neural network-based approaches are data-driven and trainable in an end-to-end fashion. 

Early QG approaches are mostly rule-based ones~\cite{heilman2009question,heilman2010good,chali2015towards}, which leverage rules or templates to generate questions. The rule-based approaches employ manually crafted rules for declarative-to-interrogative sentence transformation, typically based on syntactic~\cite{mitkov2003computer,ali2010automation,heilman2011automatic} or semantic information (Chen, 2009), while the template-based approaches generate questions using manually created templates which are predefined with placeholders to be filled with words from the source word sequence~\cite{cai2006nlgml,lindberg2013generating,song2016domain}. The major limitations of these rule-based approaches include: (1) they rely heavily on rules and templates, which are created manually and therefore expensive, (2) these rules or templates lack diversity, (3) the targets they can deal with are limited.

To tackle the limitations of rule-based approaches, the neural network-based approaches with an encoder-decoder framework are applied to the task of QG. These approaches do not rely on hand-crafted rules, and they are instead data driven and trainable in an end-to-end fashion. The release of large-scale machine reading comprehension datasets, e.g. SQuAD, MS MARCO~\cite{nguyen2016ms}, Hotpot QA~\cite{yang2018hotpotqa} and DROP~\cite{dua2019drop}, further drives the development of neural QG models. In general, these datasets contain large-scale manually annotated triples, i.e., question, answer and the context, and they can be used as training data of QG models.

As for the RNN-based models, both \newcite{du2017learning} and \newcite{yuan2017machine} apply a sequence-to-sequence model with an attention mechanism to generate questions for the text in SQuAD dataset. \newcite{zhou2017neural} enrich the RNN-based model with rich features (i.e., answer position and lexical features) to generate answer focused questions, and incorporate a copy mechanism that allows the model to copy words from the context. \newcite{duan2017question} propose to combine templates and the sequence-to-sequence model, in which they mine question patterns from a question answering community and employ a sequence-to-sequence model to generate question patterns for a given text. \newcite{tang2017question} model question answering and question generation as dual tasks, which helps generate better questions and get better question answering models at the same time.

Further, based on pointer generator network~\cite{see2017get}, \newcite{sun2018answer} propose an answer focused and position-aware model to effectively leverage answer encoding and position features.~\newcite{zhao2018paragraph} mainly focus on incorporating paragraph level context by using gated self-attention and maxout pointer networks. \newcite{nema2019let} give a refine network to mimic human process of generating questions. Besides, there has also been some work on generating questions from knowledge bases~\cite{serban2016generating,indurthi2017generating,liu2019generating}. 

More recently, pre-training models are applied to QG. For example, \newcite{dong2019unified} propose \textsc{UniLM} which makes a great progress in QG. Currently, \textsc{UniLM} is the QG model with the best performance. 

Differing from the previous work, our approach does not provide a QG model, and instead provides two extensions to the encoder-decoder framework for performance improvement. Our approach can be applied to most existing models, including RNN-based sequence-to-sequence models plus \textsc{UniLM}.

\section{Our Approach}
To begin with, we formally give the QG task definition as follows. Given the text $(x_{1}, x_{2}, ... , x_{T_{x}})$ of length $T_{x}$ as well as lexical features, i.e., named entity (NE) and part-of-speech (POS). The answer positions in this text range from $l$ to $r$. The goal is to generate a question, which is required to be as close as the reference question $(y_{1}, y_{2}, ... , y_{T_{y}})$. 

In the following subsections, we describe the details of the proposed approach to deal with the issues discussed in the previous sections. Firstly, we define a series of word transformation types, which is the basis of our approach. Secondly, we choose the pointer generator network and \textsc{UniLM} as baselines, and describe how to apply our approach by elaborating the working process of the encoder-decoder in the improved versions, respectively. 
%Thirdly, we give the computation details of the final probability distribution and the loss function.
\subsection{Word Transformation Types}

As the basis of our approach, we define the transformation types in terms of the part-of-speech of the word, as listed in Table \ref{transformation}. As for a verb, we define four types of transformation, i.e., “\#\#ing”, “\#\#vs”, “\#\#ed” and “\#\#edp”. As for a noun, we define one transformation type “\#\#ns”, which means converting the noun to its plural word. For an adjective or an adverb, we define the transformation to their comparative form as “\#\#jer” and “\#\#ver”, respectively. Similarly, we define their transformation to the superlative form as “\#\#jest” and “\#\#vest”. Note that we use the Pattern module\footnote{\url{https://github.com/clips/pattern}} in Python to conduct these transformations. As for irregular words, we manually collect a lookup table, for instance, “went” can be converted into go and “\#\#ed”.

Recall the two issues mentioned in Section 1. As for the first issue in the encoder, by a series of transformation types, we only keep the root words and the transformation types in the encoder vocabulary. As a result, it can substantially save space for other words and make the best of training data. As for the second issue, we directly copy or transform the word in the source sequence to generate questions except for the question word. A transformation type vocabulary is introduced as a necessity in the transformation type generation, which will be elaborated in the next subsections.

\begin{figure*}[t!]  
\centering  \includegraphics[width=0.70\textwidth]{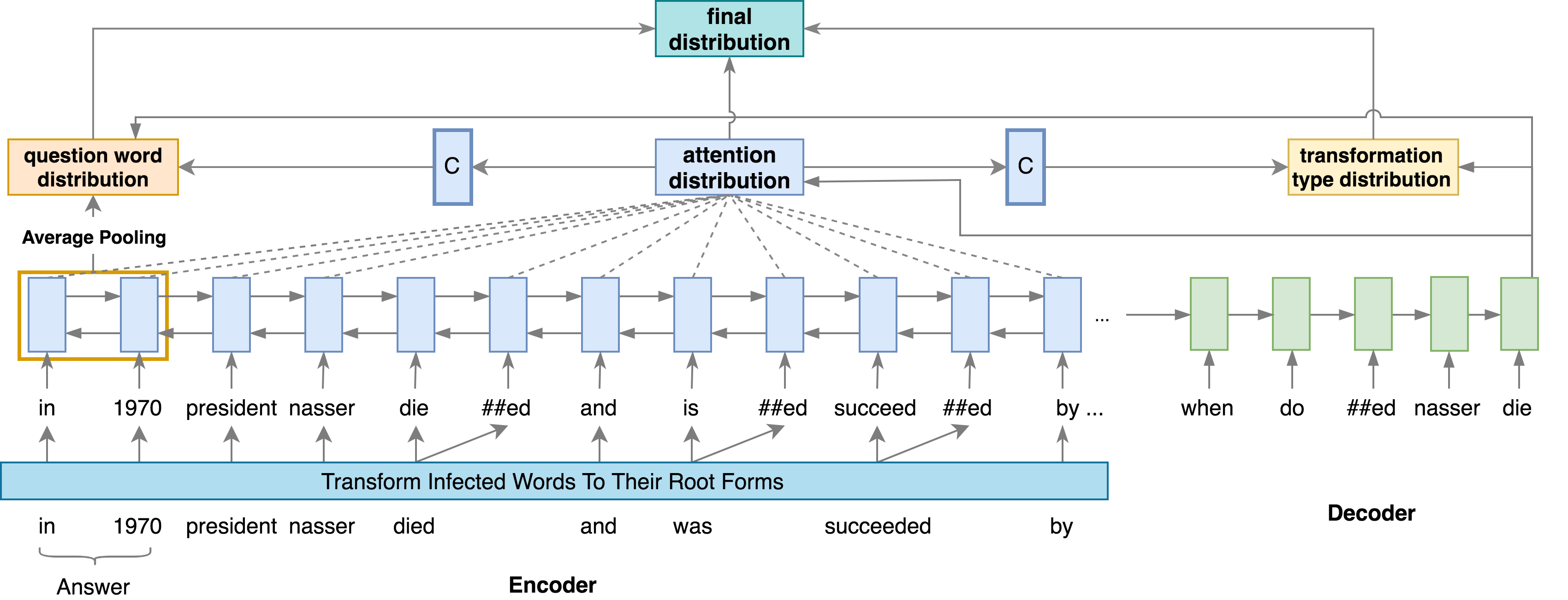} 
\vspace{-0.1cm}
\caption{The improved version based on pointer generator network. ``C" represents the context vector.}
\label{architecture}
\vspace{-0.5cm}
\end{figure*}

Before model encoding, we first get the root form of each word in the original input sequence. For example, in Figure \ref{architecture}, the word ``succeeded" is converted into its root form ``succeed" and the corresponding transformation type is ``\#\#ed". After transformation, the input sequence is converted into a sequence $(x'_{1}, x'_{2}, ... , x'_{T'_{x}})$ of length $T'_{x}$ and the answer positions range from $l'$ to $r'$. Similarly, The reference question is converted into $(y'_{1}, y'_{2}, ... , y'_{T'_{y}})$ of length $T'_{y}$. 

\subsection{Reforming Pointer Generator Network}
\subsubsection{The Encoder}

The baseline which is reformed is the attention-based pointer generator network~\cite{see2017get} enhanced with various rich features proposed by~\newcite{zhou2017neural}. These features include named entity (NE), part-of-speech (POS) and answer position in the embedding layer of the encoder.

As shown in Figure \ref{architecture}, the encoder is a bidirectional GRU, which takes as input the joint embedding of word, answer position and lexical features (NE, POS) in the form of $(w_{1}, w_{2}, ... , w_{T’_{x}})\ \text{with}\ w_i \in \mathbb{R}^{d_w+d_a+d_n+d_p}$, where $T'_{x}$ is the input length, $w_{i}$ is the embedding of $x'_{i}$ by concatenating all the feature embeddings and $d_w, d_a, d_n, d_p$ are the dimensionalities of word embedding, answer position embedding, NE embedding and POS embedding, respectively. It produces a sequence of $d_h$-dimensional hidden states $(h_{1}, h_{2}, ... , h_{T'_{x}})$, each of which is the sum of forward and backward GRU representations:

\begin{equation}
\begin{split}
h_{i} &= \overleftarrow{h}_i + \overrightarrow{h}_i, \\ 
\overleftarrow{h}_i &= \texttt{GRU}(w_i, \overleftarrow{h}_{i+1}), \\ 
\overrightarrow{h}_i &= \texttt{GRU}(w_i, \overrightarrow{h}_{i-1})
\end{split}
\end{equation}
where $\overleftarrow{h}_i, \overrightarrow{h}_i $ are all $d_h$-dimensional vectors.

\subsubsection{The Decoder}

The decoder is modified to support three actions: word copying, transformation type generation and question word generation. (1) In the word copying, the decoder generates words from the source sequence. (2) In the transformation type generation, the decoder generates from a limited transformation type vocabulary, which only includes nine types in Table \ref{transformation}. (3) In the question word generation, the decoder generates question words from a restricted question word vocabulary. 

%\noindent
%\textbf{}
\noindent \textbf{Word Copying}
The decoder is a unidirectional GRU conditionally taking all the encoded hidden states as input. At decoding step $t$, the decoder reads an input word embedding $w_{t}$, previous attentional context vector $c_{t-1}$ and its previous hidden state $s_{t-1}$ to update its current hidden state $s_{t} \in \mathbb{R}^{d_h}$:
\begin{equation}
s_{t} = \texttt{GRU}([w_{t}; c_{t-1}], s_{t-1})
\end{equation}

The context vector $c_t$ is generated through an attention mechanism~\cite{bahdanau2014neural}. At time step $t$, the context vector $c_t$ is calculated as follows:
%the attention distribution $\alpha_t$ are
\begin{equation}
\begin{split}
&c_t = \sum_{i=1}^{T'_{x}}\alpha_{ti}h_{i} \\
\alpha_{ti} &= \texttt{Softmax}(e_{ti}) \\
e_{ti} = v^{T}&\tanh (W_{h}^T h_{i} + W_{s}^T s_{t} + b_{})
\end{split}
\end{equation}
where $\alpha$ is the attention distribution and $W_{h}, W_s, b$ and $v$ are all trainable parameters.

The attention distribution can be viewed as a semantic matching between hidden states of the encoder and the hidden state of the decoder. It indicates how the decoder cares about different hidden states of the encoder during decoding. Therefore, the attention distribution can be regarded as the probability distribution of the word copying.
%spreads out the amount it cares about different encoder hidden states during decoding
\begin{equation}
P_{copy} = \alpha_{t}
\end{equation}

\noindent \textbf{Transformation Type Generation}
Before encoding, we have already converted each word to its root form, i.e., the root word. In the decoding, besides coping words, we should endow the model the ability to transform some words to their proper forms, e.g., the ``do" in Figure \ref{architecture} is generated from question word vocabulary and we need to transform it to ``did". To carry out this transformation, we introduce the transformation type generation and generate the transformation type for the root word, e.g., ``\#\#ed" is generated to transform ``do" to ``did".

%We define this mode to transform the word from source sequence in the process of generating questions. All the transformation types are defined above. 
We get the transformation type distribution by the following function.

\begin{equation}
P_{trans} = \texttt{Softmax}\left(g_1(s_{t}, c_t)\right)
\end{equation}
where $g_1(\cdot)$ is a two-layer feedforward neural network with a maxout internal activation. $P_{trans} \in \mathbb{R}^{|V_{trans}|}$ denotes the probability distribution of transformation types.
%with a vocabulary size of $|V_{trans\_type}|$.

\noindent \textbf{Question Word Generation} We find that the generation of question words is mainly determined by the answer and its surrounding words. For example, in Figure \ref{architecture}, the answer and its context ``in 1970" already involve the essential information to generate the question word ``when", which suggests that the answer and its surrounding words can benefit the question word generation. We also find that the total number of question words is limited. Therefore, we introduce a specific vocabulary of question words to directly and explicitly model the question words generation. Note that the question word vocabulary contains top-$1000$ frequent words in reference questions of the training set, which means that it also contains other question-related words.

As depicted in Figure \ref{architecture}, in the question word generation, the model generates question words based on a restricted vocabulary of question words. This action produces a question word distribution based on an answer embedding $v_{answer}$, the decoder state $s_{t}$ and the context vector $c_t$:
\begin{equation}
P_{quest}= \texttt{Softmax} \left(g_2(v_{answer}, s_{t}, c_t)\right)
\end{equation}
where $g_2(\cdot)$ is a two-layer feedforward neural network, $P_{quest}$ is a $|V_{quest}|$-dimensional probability distribution, and $|V_{quest}|$ is the size of vocabulary of question words. We employ the average pooling function to calculate the answer embedding.
\begin{equation}
v_{answer} = \frac{\sum_{t=l'}^{r'}h_{t}}{r'-l'+1}
\end{equation}
%hidden state at the answer start position as the answer embedding, i.e., $v_{ans} = h_{answer\_start}$. We argue that under bidirectional encoding, this answer embedding has already memorized both the left and the right contexts around the answer region, making it a desired choice.

\noindent \textbf{Three-action Combination}
To control the balance among different actions, we introduce a three-dimensional switch probability, acting as a three-way soft switch:
\begin{equation}
p_{quest},p_{copy},p_{trans} = \texttt{Softmax}(f(c_t,\! s_{t},\! w_{t}))
\end{equation}
where $f(\cdot)$ is a one layer feedforward network. We compute the final probability distribution through a weighted summation of the three action probability distributions:
\begin{equation}\label{eq:11}
\begin{split}
P(w) &= p_{copy} P_{copy}(w) + p_{trans} P_{trans}(w) \\
     &+ p_{quest} P_{quest}(w)
\end{split}
\end{equation}

\subsection{Reforming \textsc{UniLM}}
\textsc{UniLM} is a pre-trained language model which has the same structure as BERT\cite{devlin-etal-2019-bert} but can be used for natural language generation tasks. The model is firstly pre-trained on a large scale corpus which enables it to learn flexible language representations. Then, it is fine-tuned on the dataset of the downstream tasks using the sequence-to-sequence language model objective. In \textsc{UniLM}, there is no explicit separation between the encoder and decoder. In contrast, the source and target sequence are fed to the model together. The sequence-to-sequence generation is implemented by a special self-attention mask $M$. 

%\subsubsection{Sequence-to-Sequence Language Model}
Following~\newcite{dong2019unified}, we concatenate the answer segment $(x'_{l'},...,x'_{r'})$ and the text segment $(x'_1,x'_2,...,x'_{T'_x})$ forming the new source segment $X=(x^*_1,...,x^*_{T^*_x})=(x'_1,x'_2,...,x'_{T'_x},\text{[EOS]},x'_{l'},...,x'_{r'})$. Next, we concatenate the new source segment and the target segment $Y=(y'_{1}, y'_{2}, ... , y'_{T'_{y}})$ as input, i.e., $\text{[SOS]}X\text{[EOS]}Y\text{[EOS]}$. Then we feed the input to the model. In training, we randomly replace tokens with $\text{[MASK]}$ in $Y$ at a rate of 0.8.

\begin{figure}
    \centering
    \includegraphics[width=0.5\textwidth]{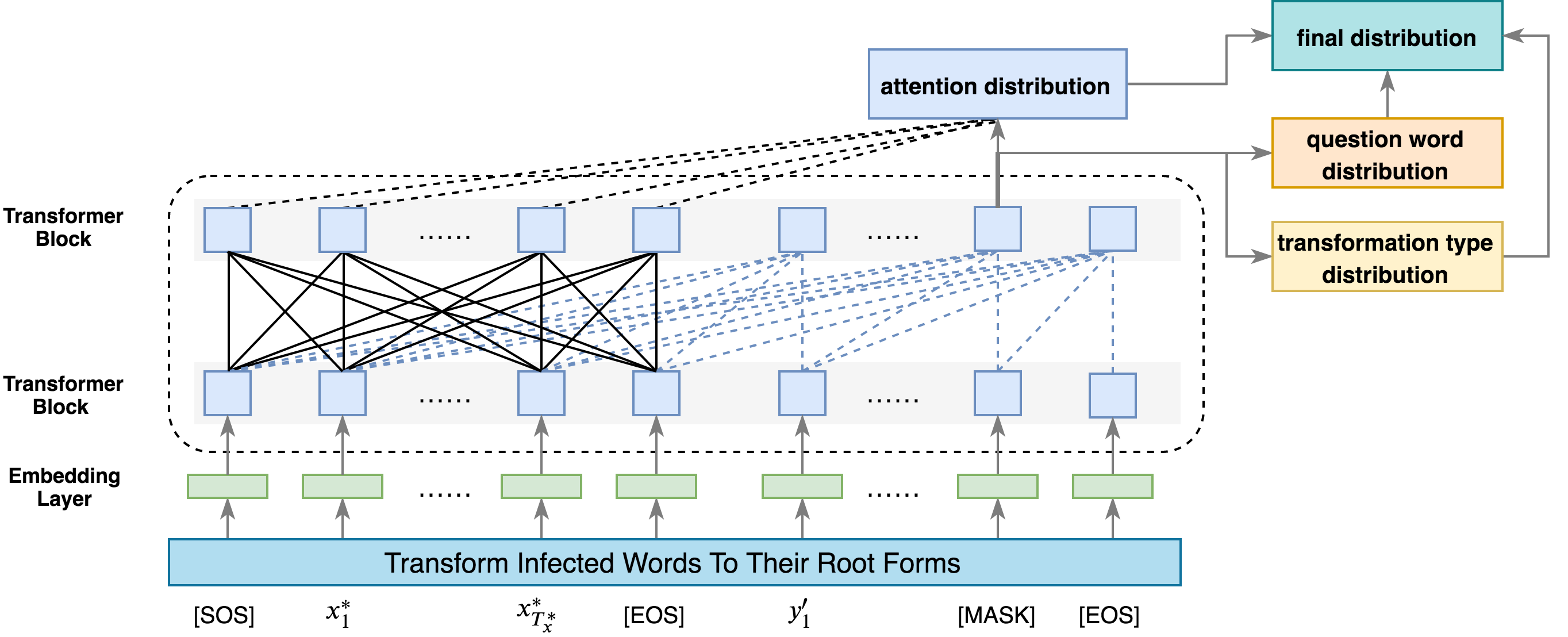}
    \vspace{-0.1cm}
    \caption{Model architecture of the improved version based on \textsc{UniLM}. Note that there should be L transformer layers. For simplicity, we only keep two of them in this figure.}
    \label{fig:unilm}
    \vspace{-0.5cm}
\end{figure}

For each input token, its embedding is the sum of the token embedding, position embedding and segment embedding. As shown in Figure \ref{fig:unilm}, we get token embeddings $w=(w_{1}, w_{2}, ... , w_{T^*_{x}+T'_{y}+3})$, which is then feed into $L$-layer transformers to get the deep representation $H^l = h_{i}^l, i \in [1, T^*_{x}+T'_{y}+3]$.

\begin{equation}
h_{i}^{l} = \texttt{Transformer}(h_{i}^{l-1}), l = 1,2,3,...,L
\end{equation}
where $H^{0}$ is initialized with $w$. The output of the self-attention head of the $l$-th transformer is calculated by:
\begin{equation}
Q_l = H^{l-1}W^{Q},~K_l = H^{l-1}W^K,~V_l = H^{l-1}W^V
\end{equation}
\begin{equation}
A_l = \texttt{Softmax}(\frac{Q_lK_l^T}{\sqrt{d_k}} + M)V_l
\end{equation}
where $M \in \mathbb{R}^{(T^*_{x}+T'_{y}+3) \times (T^*_{x}+T'_{y}+3)}$. $M_{ij} = 0$ means it is allowed to attend, while $M_{ij} = -\infty$ indicates it is not allowed to attend. We can find that the special self-attention mask $M$ allows the token to be generated to attend to all of the source tokens and the preceding generated tokens. This mechanism allows the model to generate text in a sequence-to-sequence fashion. 

%The model is designed to be a sequence-to-sequence language model by predicting the masked tokens with the restrictions that tokens in $Y$ can only attend to its previous tokens. As a result, it trains a unidirectional decoder. For more details, please refer to ~\newcite{dong2019unified}.

%\subsubsection{The Decoding Process}
%During decoding, we feed a [MASK] token to the model at each step. The model firstly send the embedding to the L-layer Transformer structure which performs self-attention with all of the preceding hidden states including the input tokens and the generated tokens. 
To adopt the word transformation approach, we directly compute the distributions associated with the three actions based on the hidden state of the current step. The details are shown as follows.
% on the \textsc{UniLM}

\noindent \textbf{Word Copying}
The attention distribution over the source tokens is obtained by an extra attention module which computes the attention scores using dot product:
\begin{gather}
    P_{copy} = \texttt{Softmax}(g_3(h_t)) \\
    % P_{copy,i} = \frac{\exp{(e_{ti})}}{\sum_{j}^{T'_x}\exp{(e_{tj})}} \\
    g_3(h_t)=\left[g_3(h_t)_i\right]_{i=1}^{T'_x}=\left[h_t \cdot h_i\right]_{i=1}^{T'_x}
    %g_{3,i}(h_t) = h_t \cdot h_i \quad (i\in[1,T'_x])
\end{gather}
where $h_t$ is the hidden state of $t$-th token output by the last layer of transformer. The output of $g_3(h_t)$ is the dot product between $h_t$ and hidden states of all of the input tokens.

\noindent \textbf{Transformation Type Generation}
The transformation type distribution is obtained by computing the dot product between the current hidden state and the embeddings of the transformation types. 
\begin{gather}
    P_{trans} = \texttt{Softmax}({g_4(h_t)}) \\
    % P_{trans_type} = \frac{\exp{(a_{ti})}}{\sum_{j}^{|V_{trans\_type}|}{\exp{(a_{tj})}}} \\
    g_4(h_t)=\left[g_4(h_t)_i\right]_{i=1}^{|V_{trans}|}=\left[h_t \cdot e^T_i\right]_{i=1}^{|V_{trans}|}
    %g_{4,i}(h_t)=h_t \cdot e^T_i \quad (i\in\left[1,|V_{trans\_type}|\right])
\end{gather}
where $e^T$ is the embedding of the transformation types. $g_4(h_t)$ is the dot product between $h_t$ and the embeddings of all of the transformation types. $P_{trans}$ denotes the probability distribution of transformation types.
%Same as the reformed pointer generator network, 

\noindent \textbf{Question Word Generation} 
Similar to the calculation of the distribution of the transformation types, the distribution of the question words is obtained by the dot product between the current hidden state and the embeddings of the question words. 
\begin{gather}
     P_{quest} = \texttt{Softmax}({g_5(h_t)}) \\
    % P_{trans_type} = \frac{\exp{(a_{ti})}}{\sum_{j}^{|V_{trans\_type}|}{\exp{(a_{tj})}}} \\
    g_5(h_t)=\left[g_5(h_t)_i\right]_{i=1}^{|V_{quest}|}=\left[h_t \cdot e^Q_i\right]_{i=1}^{|V_{quest}|}
    %g_{5,i}(h_t)=h_t \cdot e^Q_i \quad (i\in\left[1,|V_{quest\_word}|\right])
\end{gather}
where $e^Q$ is the embedding of the question words. $g_5(h_t)$ is the dot product between $h_t$ and the embeddings of all of the question words $e^Q$.

\noindent \textbf{Three-action Combination}
The three-dimensional switch probability is obtained by the current hidden state $h_t$.
\begin{equation}
p_{quest},p_{copy},p_{trans} = \texttt{Softmax}(f_1(h_t))
\end{equation}
where $f_1(\cdot)$ is a one layer feedforward network. The final probability distribution is computed as the same as Equation \ref{eq:11}.
\section{Experiments}
\begin{table*}[!th]
\centering
\subfloat[]{%
\scalebox{0.85}{
\begin{tabular}{lcccccc}
%\hline \bf DataSet & \multicolumn{6}{c}{\bf SQuAD} \\
\hline\bf Model & \bf BLEU1 & \bf BLEU2 & \bf BLEU3 & \bf BLEU4 & \bf ROUGE-L & \bf METEOR \\ \hline
%NQG++ \cite{zhou2017neural} & 42.36 & 26.33 & 18.46 & 13.51 & 41.60 & 18.18 & 43.93 & 31.25 & 22.97 & 17.54 & 45.35 & 19.19\\
%PG (Pointer Generator) \newcite{see2017get} & 32.32 & 18.04 & 12.06 & 8.60 & - & - & 42.40 & 29.37 & 20.71 & 15.16 & - & - \\
Pointer & 40.49 & 26.11 & 18.94 & 14.34 & 42.15 & 18.71 \\
%AFPA \cite{sun2018answer} & 43.02 & 28.14 & 20.51 & 15.64 & 43.32 & 19.65 & 48.24 & 35.95 & 25.79 & 19.45 & 47.11 & 21.24 \\
%RefNet  & 47.27 & 31.88 & 23.65 & 18.16 & - & - & - & - \\
\textbf{Pointer + WT} & \bf 45.08 & \bf 29.56 & \bf 21.54 & \bf 16.41 & \bf 45.40 & \bf 20.61 \\
% &  +4.59 &  +3.45 &  +2.60 &  +2.07 &  +3.25 &  +1.90 \\
\hline
\hline
\textsc{UniLM} & 49.82 & 34.48 & 26.03 & 20.39 & 49.02 & 23.52 \\
\textbf{\textsc{UniLM} + WT} & \bf 51.56 & \bf 35.78 & \bf 27.06 & \bf 21.24 & \bf 50.46 & \bf 23.93 \\
% & +1.74 & +1.30 & +1.03 & +0.85 & +1.44 & +0.41\\
\hline
\end{tabular}
}}%
\vspace{-0.2cm}

\subfloat[]{%
\scalebox{0.85}{
\begin{tabular}{lcccccc}
%\hline \bf DataSet & \multicolumn{6}{c}{\bf MARCO} \\
\hline\bf Model & \bf BLEU1 & \bf BLEU2 & \bf BLEU3 & \bf BLEU4 & \bf ROUGE-L & \bf METEOR \\ \hline
%NQG++ \cite{zhou2017neural} & 42.36 & 26.33 & 18.46 & 13.51 & 41.60 & 18.18 & 43.93 & 31.25 & 22.97 & 17.54 & 45.35 & 19.19\\
%PG (Pointer Generator) \newcite{see2017get} & 32.32 & 18.04 & 12.06 & 8.60 & - & - & 42.40 & 29.37 & 20.71 & 15.16 & - & - \\
Pointer & 44.45 & 31.85 & 23.32 & 17.90 & 46.07 & 20.02\\
%AFPA \cite{sun2018answer} & 43.02 & 28.14 & 20.51 & 15.64 & 43.32 & 19.65 & 48.24 & 35.95 & 25.79 & 19.45 & 47.11 & 21.24 \\
%RefNet  & 47.27 & 31.88 & 23.65 & 18.16 & - & - & - & - \\
\textbf{Pointer + WT} & \bf 56.14 & \bf  39.36 & \bf 29.04 & \bf 22.10 & \bf 59.29 & \bf 26.40\\
% &  +4.59 &  +3.45 &  +2.60 &  +2.07 &  +3.25 &  +1.90 & +11.69 &  +7.51 & +5.72 & +4.20 & +13.22 & +6.38\\
\hline
\hline
\textsc{UniLM} & 60.57 &  43.14 & 32.38 & 25.01 & 60.58 & 29.31 \\
\textbf{\textsc{UniLM} + WT} & \bf 62.65 & \bf  45.33 & \bf 34.31 & \bf 26.55 & \bf 62.85 & \bf 29.72\\
% & +1.74 & +1.30 & +1.03 & +0.85 & +1.44 & +0.41 & +2.08 &  +2.19 & +1.93 & +1.54 & +2.27 & +0.41\\
\hline
\end{tabular}
}}%
\vspace{-0.4cm}
\label{result}
\caption{The main experimental results of baselines and improved versions on SQuAD (a) and MARCO (b). ``WT" means the proposed word transformation approach. }
%Relevance scores and document rankings of the case in Figure \ref{fig:1}
\label{result}
\vspace{-0.3cm}
%\vspace{-0cm}
\vspace{-0cm}
\end{table*}
\subsection{Experiment Settings}
\textbf{Dataset} We conduct the experiments on SQuAD and MARCO datasets. Since the test sets of these datasets are not publicly available, we follow \newcite{zhou2017neural} to randomly split the development set into two parts and use them as the development set and test set for the QG task. In SQuAD, there are $86,635$, $8,965$ and $8,964$ question-answer pairs in our training set, development set and test set, respectively. We directly use the extracted features\footnote{\url{https://res.qyzhou.me/redistribute.zip}} shared by \newcite{zhou2017neural}. In MARCO, there are $74,097$, $4,539$ and $4,539$ question-answer pairs in our training set, development set and test set, respectively. We use Stanford CoreNLP\footnote{\url{https://nlp.stanford.edu/software/}} to extract lexical features. 
%We attach all the processed data sets in the supplemental materials. 

\noindent
\textbf{Implementation Details}
We set the cutoff length of the input sequence as 128 words. The encoder vocabulary contains the most frequent $30,000$ words in each training set. The decoder vocabulary contains two sub-vocabularies. One is the question word vocabulary which contains most frequent $1000$ words in the reference questions of the training set. The other one is the transformation type vocabulary, including 9 transformation types as described in Section 3. For the RNN-based model, we use the pre-trained Glove word vectors\footnote{\url{http://nlp.stanford.edu/data}} with 300 dimensions to initialize the word embeddings that will be further fine-tuned in the training stage. The representations of answer position feature and lexical features at the embedding layer of the encoder are randomly initialized to 32 dimensional vectors that are trainable during training stage. The size of hidden states of both the encoder and decoder is $512$. We use dropout only in the encoder with a dropout rate $0.20$. The size of answer embedding is $512$. We use the optimization algorithm Adam~\cite{kingma2014adam} with the learning rate $0.002$ and we set the batch size as $32$. As for the \textsc{UniLM}, we adopt the base version of the \textsc{UniLM} and use the recommended parameters as detailed by \newcite{dong2019unified}. After training, we select the best model on the development set for testing. 
%We use gradient clipping with a maximum gradient norm of $2$

\noindent
\textbf{Evaluation Metrics} We evaluate our approach using n-gram similarity metrics, i.e., BLEU~\cite{papineni2002bleu}, ROUGE-L~\cite{lin2004rouge} and METEOR~\cite{lavie2009meteor}.

\noindent
\textbf{Competitors} In the experiments, we have the following four competitors for comparisons, where Pointer and \textsc{UniLM} are baselines and the other two are corresponding improved versions.
\begin{itemize}
    \setlength{\itemsep}{2pt}
    \setlength{\parsep}{0pt}
    \setlength{\parskip}{0pt}
    %\item \textbf{NQG++~\cite{zhou2017neural}} It is a RNN-based neural question generation system on SQuAD that incorporates rich features to the embedding layer of a sequence-to-sequence model and introduces copy mechanism proposed by ~\newcite{gulcehre2016pointing}.
    \item \textbf{Pointer generator network (Pointer)} It is a typical RNN-based sequence-to-sequence model with the copy mechanism~\cite{see2017get}. To make a fair comparison, the lexical features are added to the embedding layer as same as~\newcite{zhou2017neural}.
    %\item \textbf{Feature-enriched pointer-generator network} We add the features to the embedding layer of the pointer-generator model as described in \cite{zhou2017neural}.
    %\item \textbf{Answer-Focused and Position-Aware (AFPA) model} We implement the model reported in \cite{sun2018answer}, which is a previous state-of-the-art model based on pointer generator model.
    \item \textbf{Pointer generator network plus Word Transformation approach (Pointer + WT)} We enhance the Pointer model with the proposed word transformation approach.
    \item \textbf{\textsc{UniLM}~\cite{dong2019unified}} It is a pre-trained language model which can be applied to natural language generation tasks. 
    \item \textbf{\textsc{UniLM} + WT} We combine the \textsc{UniLM} model with the proposed word transformation approach.
\end{itemize}

Pointer and Pointer + WT are implemented with Tensorflow, while \textsc{UniLM} and \textsc{UniLM} + WT are implemented by PyTorch. 

\noindent
\subsection{Performance Evaluation}
Table \ref{result} shows the main results, and we have the following observations:
\begin{itemize}
    %\item The pointer-generator model outperforms NQG++. Both of the two models employ the sequence-to-sequence model with copy mechanism and the same features. The major difference between them is that their copy mechanism has different architecture, and pointer-generator shows better performance.
    %\item The pointer-generator model without the features does not perform well. This verifies the effectiveness of the features extracted by ~\newcite{zhou2017neural}.
    %\item The answer-focused model and position-aware model outperform the feature-enriched pointer-generator model and NQG++. 
    \setlength{\itemsep}{2pt}
    \setlength{\parsep}{0pt}
    \setlength{\parskip}{0pt}
    
    \item The Pointer + WT model performs better than the original Pointer model which indicates that generating questions by word transformation is effective.
    \item \textsc{UniLM} + WT model can still significantly outperform the powerful \textsc{UniLM}, which indicates the effectiveness of applying our approach to the pre-trained language models for QG tasks.
    \item Our approach makes greater improvement over MARCO than SQuAD. That might be because MARCO is extracted from the search engine, which induces a bias over word copying and word transformation generation from source text.
\end{itemize}

%\begin{figure*}[h]  
%\centering  \includegraphics[width=1.0\textwidth]{time_complexity.png} 
%\caption{ Time consuming of one step decoding in different models. Note that we regard the decoding of root word and its transformation type as one step in efficiency comparison.}
%\label{architecture}
%\end{figure*}

%Time consuming of one step decoding in different models. Note that we regard the decoding of root word and its transformation type as one step in efficiency comparison.

Besides generation quality, we also compare baselines with the corresponding improved versions on efficiency of decoding. We record the time of generating a word of a question given the test text with a beam size of 12 and calculate the average value. Note that we regard the decoding of root word and its corresponding transformation type as one word in efficiency comparison. The comparison is conducted on a GPU environment with a single Tesla V100. For the RNN-based models, the Pointer + WT model is more efficient which can save more than \textbf{39\%} (from 0.0081s to 0.0049s) decoding time compared to original model. For the pre-trained models, the \textsc{UniLM} + WT model can save \textbf{13\%} decoding time (from 0.0144s to 0.0125s). We also find that \textsc{UniLM} behaves better than Pointer on generation quality, but it sacrifices efficiency. Specifically, it is nearly one time slower than the RNN-based models. From the comparisons on both effectiveness and efficiency, we can conclude that this word transformation approach can enhance the quality of QG and speed up the decoding at the same time.

%\subsection{Human Evaluation}
%We also conduct human evaluation to examine the quality of the questions generated by the models and reference questions by scoring them on a scale of 0 to 3 in terms of fluency, answer related and . As reported in Table 8, our model generates questions with higher scores on the three metrics than the two baseline models, indicating the superiority of our proposed model by utilizing the sentence-level semantics and answer position-awareness
\subsection{Human Evaluation}
We further conduct human evaluation to analyze the generation quality of the Pointer model and the Pointer + WT model. We randomly sample 200 cases from the SQuAD dataset and ask three annotators specialized in language to compare the generation quality. The annotators are shown two questions, one generated by the Pointer model and the other one by the Pointer + WT model. They are asked which one is better by three factors, i.e., fluency, completeness, and answerability. These three factors are equally important in this evaluation. The final label is determined by majority voting. By the metric value, we get that the Pointer + WT model outperforms the Pointer model. Specifically, in \textbf{73/200} cases, the Pointer + WT model is better compared to the Pointer model. In 91/200 cases, the two models behave nearly the same. In 36/200 cases, the Pointer + WT model behaves worse than the baseline.

\begin{table}[t]
\centering
\begin{tabular}{p{0.9\columnwidth}}
\tabucline[1pt]{-}
	\textbf{Text:} In the March Battle of Fort Bull, French forces destroyed the fort and large quantities of supplies , including \underline{\textbf{45,000}} pounds of gunpowder . \\
\textbf{Answer:} \textbf{45,000} \\
\textbf{Reference question:} How much gun powder was destroyed in attack ? \\
\hline
\hline 
\textbf{Pointer:} 
How much of gunpowder \textbf{\color{cornellred}{'s}} were \textbf{\color{cornellred}{killed}} in the March Battle of Fort Bull ?  \\
\textbf{Pointer + WT:} 
How much gunpowder was destroyed in the March Battle of Fort Bull ? \\
\tabucline[1pt]{-}
\end{tabular}
%\vspace{-0.1cm}
\caption{A case where the Pointer model generates unrelated words, i.e., ``\textbf{killed}" and ``\textbf{'s}", while the model incorporating word transformation approach can generate the question more correctly.} 
\label{tab:case}
\vspace{-0.3cm}
\end{table}

Table \ref{tab:case} gives a case to demonstrate the effectiveness of our approach. As shown in Table \ref{tab:case}, the Pointer model generates unrelated words, i.e., ``\textbf{killed}" and ``\textbf{'s}". It might be because the Pointer model generates from a large and noisy vocabulary. However, the Pointer + WT model can generate the question more correctly, which can be definitely owed to our approach.

%In order to verify that the word transformation method improves the effectiveness as we think. We conduct a case study where the Pointer + WT performs better than Pointer model. As shown in Table \ref{tab:case}, because Pointer model generates from a large and noisy vocabulary, it generates unrelated word, i.e.,``ban". However, by fusing word transformation the model can generate more correctly.

\section{Conclusion}
In this paper, we discover two major issues in the existing neural QG models. To tackle the two issues, we propose this enhancing approach for QG and apply the approach to two typical sequence-to-sequence models, i.e., the pointer generator network and \textsc{UniLM}. We further conduct extensive experiments using SQuAD and MARCO datasets. The experimental results show that improved versions of models can significantly enhance the quality of QG and speed up the decoding.

\newpage
%\section{Limitations}
%In this paper, we utilized the benefits of rule-based question generation method and sequence-to-sequence models. The proposed method is applicable in real generation applications for the purpose of more accurate and controllable generation. However, there still exist some limitations. (1) The model is not as efficient as rule-based methods. It is still computational complex to employ it online. (2) The model generates from a limited vocabulary which restricts the diversities of generated questions. 

\bibliography{anthology,acl2020}
\bibliographystyle{acl_natbib}

\end{document}